%% file: Run_This_Example.tex
\documentclass[Alon2,ChapterTOCs,singlecolor,11pt,pdf,oneside,openany]{Alon}
\let\nobreak\relax
\usepackage{fixltx2e,fix-cm}
\usepackage[english]{babel}
\usepackage[T1]{fontenc}
\usepackage{lmodern}
\usepackage{amssymb}
\usepackage{amsmath,amsthm}

\usepackage{graphicx}
\graphicspath{{chapter1/figures/}{figures/}}

\usepackage{booktabs}
\usepackage{multirow}
\usepackage{makecell}
\usepackage{array}
\usepackage{adjustbox}
\usepackage{tabularx}
\usepackage{algorithm}
\usepackage{algorithmic}
\usepackage{enumitem}
\usepackage{makeidx}
\usepackage{multicol}
\usepackage{changepage}
\usepackage{version}
\usepackage{lipsum}
\usepackage{epstopdf}
\usepackage[caption=false]{subfig}
\usepackage{placeins}

\renewcommand{\arraystretch}{1.15}

\makeindex

\usepackage{tikz,xcolor}
\usepackage[hypertexnames=false]{hyperref}
\hypersetup{
  hidelinks,
  pdfborder={0 0 0}
}
\makeatletter
\providecommand\Hy@tocdestname{}
\makeatother
\definecolor{lime}{HTML}{A6CE39}
\DeclareRobustCommand{\orcidicon}{%
	\begin{tikzpicture}
	\draw[lime, fill=lime] (0,0)
	circle [radius=0.16]
	node[white] {{\fontfamily{qag}\selectfont \tiny ID}};
	\draw[white, fill=white] (-0.0625,0.095)
	circle [radius=0.007];
	\end{tikzpicture}
	\hspace{-2mm}
}

\foreach \x in {A, ..., Z}{%
	\expandafter\xdef\csname orcid\x\endcsname{\noexpand\href{https://orcid.org/\csname orcidauthor\x\endcsname}{\noexpand\orcidicon}}
}

\renewcommand\bibfont{\fontsize{10}{12}\selectfont}

\theoremstyle{plain}

\theoremstyle{definition}

\theoremstyle{remark}

\title{Taylor and Francis Book Chapter}

\begin{document}

\frontmatter

\maketitle

\tableofcontents

\mainmatter

\include{chapter1/ch1}

\end{document}

%% file: chapter1/ch1.tex
\chapterauthor{Md. Maruf Bangabashi}{Department of Computer Science and Engineering, Dhaka International University, Dhaka-1212, Bangladesh.}
\chapterauthor{Tahmid Hasan}{Department of Computer Science and Engineering, Bangladesh University of Engineering and Technology, Dhaka-1000, Bangladesh.}
\chapterauthor{Golam Mahmud}{Department of Computer Science and Engineering, Dhaka International University, Dhaka-1212, Bangladesh.}
\chapterauthor{Md. Mostafijur Rahman}{Department of Computer Science and Engineering, Dhaka International University, Dhaka-1212, Bangladesh.}
\chapterauthor{Md. Toufiqur Rahman}{Department of Computer Science and Engineering, Dhaka International University, Dhaka-1212, Bangladesh.}
\chapterauthor{Jahanur Biswas*}{Department of Computer Science and Engineering, Dhaka International University, Dhaka-1212, Bangladesh.}

\chapter{Echoes of Unrest: A Multimodal NLP Framework for Early Warning of Fake News and Violence-Driven Mob Activity}

\chaptermark{Echoes of Unrest: Multimodal NLP for Early Warning}

\chapterinitial{R}apid growth in social media has transformed global communication by enabling fast information exchange, but it has also accelerated the spread of misinformation. Fake news, manipulated content, and provocative narratives are increasingly linked to social unrest, political instability, and mob violence. Incidents in South Asia and elsewhere demonstrate how false information disseminated via platforms such as Facebook and WhatsApp can trigger real-world harm, often spreading faster than fact-checking efforts can respond. To address this challenge, this chapter presents a multilingual, multimodal Natural Language Processing (NLP) framework for early detection of misinformation and violence-prone dynamics. A fused dataset of 138,256 Bangla and English samples was created by combining multiple benchmark datasets. The framework integrates XLM-RoBERTa for multilingual text representation, CLIP for visual embedding, and a multi-head attention mechanism for multimodal fusion, enhanced with auxiliary features such as sarcasm and geospatial metadata. Experiments on a stratified 30\% subset achieved 98\% test accuracy with strong precision and recall. The outcomes show the efficacy of multimodal approaches in early misinformation detection and highlight the added value of geospatial signals for anticipating real-world escalation. 

\section{Introduction}
Social media usage has grown exponentially and changed the way people communicate around the world. They make the exchange of information fast and provide a means of sharing misleading information. In recent years, false news and photographs, sarcastic or provocative comments have been associated with social unrest, political instability, and mob violence in many countries. The dangers of online misinformation are evident in incidents of mob lynchings in South Asia, fueled by false content spread via Facebook and WhatsApp. Similar crises have happened in most regions across the world, where the information that spreads is more likely than the verification procedures.

Current fake news detection systems have made considerable progress through text-based natural language processing methods. However, these systems remain limited in three important ways. First, they are largely language-focused and rely primarily on textual data, thereby overlooking low-resource languages such as Bangla \cite{singh2021detecting}. Second, they frequently fail to capture multimodal misinformation that combines text with images and stylistic cues to strengthen deceptive claims \cite{alam2021survey}. Third, they generally do not consider geospatial factors, which restrict the identification of localized misinformation hotspots that may provoke mob actions \cite{tufchi2023comprehensive}.

Recent studies have shown the effectiveness of multimodal misinformation detection frameworks \cite{singh2021detecting,alam2021survey,tufchi2023comprehensive,mostafa2024modality}. Nevertheless, no unified framework adequately combines multilinguality, multimodality, sarcasm awareness, and geospatial signals for violence monitoring. This chapter therefore addresses the following question: how can a multilingual, multimodal NLP framework be developed to accurately detect fake news and misinformation across diverse languages, modalities, and geospatial contexts while also providing early warning for the escalation of misinformation into violence-driven mob activity?

The proposed framework addresses this gap by integrating multilingual text, visual cues, stylistic signals such as sarcasm, and geospatial metadata into a unified misinformation detection system. It is particularly relevant for low-resource language contexts, including Bangla, where misinformation can have serious social consequences but remains understudied in existing systems.

Geospatial visualizations of misinformation hotspots can be valuable for policymakers by helping them identify regions at risk of violence or unrest and enabling proactive intervention. The system can also help security officials allocate resources more effectively on the basis of the geographical distribution of misinformation. Fact-checkers and journalists may benefit from real-time misinformation warnings that support timely verification before publication. As a result, the proposed framework provides a real-time, automated misinformation detection mechanism that combines multimodal analysis with geospatial awareness and thus supports informed decision-making and public safety.

This chapter's main contributions are as follows:
\begin{itemize}[itemsep=0ex]
\item A heterogeneous fused dataset of more than 138,000 records is constructed by integrating four resources: Bangla and English misinformation-related resources with differing modality coverage. 
\item An attention-based multimodal fusion model is introduced to jointly integrate
textual, visual, stylistic, and geospatial features for misinformation detection.
\item The proposed framework achieves an accuracy of 98\%, demonstrating strong predictive performance on the fused corpus.
\end{itemize}

This is how the rest of the chapter is structured. In Section~2, the suggested framework is placed in context, and relevant work is reviewed. The materials and techniques, such as system design and dataset construction, are described in Section~3. The experimental setup and evaluation findings are shown in Section~4. The chapter is finally concluded, and future research possibilities are outlined in Section~5.

\section{Related Work}
Several research studies have been done on the identification of fake news and misinformation, taking into account the increasing prevalence of social media sites. However, the initial approaches focused mainly on the use of text data. Specifically, Singh et al. \cite{singh2021detecting} utilized text data to detect fake news written in English. Despite contributing positively to the field, these studies cannot be considered comprehensive, since they neglect other useful data sources, including images and geographic metadata.

\subsection{Text-Based Approaches}
The conventional approaches employed in the identification of misinformation content were based largely on text-based characteristics like linguistics, sentiments, and lexical aspects. In their research, Singh et al. \cite{singh2021detecting} applied linguistic and sentiment cues to determine any misleading information. Similarly, Segura-Bedmar and Alonso-Bartolome \cite{segura2022multimodal} suggested a system that considered text-based characteristics to detect disinformation. Though the techniques may be quite efficient in some situations, their weaknesses cannot be overlooked. According to Alam et al. \cite{alam2021survey}, these techniques often overlook significant visual characteristics and cannot perform efficiently in instances where sarcasm is present.

\subsection{Multimodal Approaches}
A number of studies in recent years have explored multimodal solutions for the detection of fake news. In Qi et al. \cite{qi2021improving} an entity-centric multimodal approach to integrate text and visual data was offered to improve detection outcomes. Also, Xu et al. \cite{xu2025mdam3} offered MDAM3, a deep learning method to integrate text, images, and videos. Even though these methods represent significant achievements in terms of fake news detection, they require high computational power and tend to ignore geospatial attributes. This becomes especially important because of potential misinformation leading to violence in certain regions.

Moreover, Bansal et al. \cite{bansal2024mmcfnd} presented an approach to caption-aware multimodal analysis of low-resource Indic languages. On the other hand, Sharma and Arya \cite{sharma2024mmhfnd} developed a multimodal multiclass approach to detecting Hindi fake news using contrastive learning. However, even though this development represents significant progress, its practical applicability is limited due to language specificity and a lack of geospatial information about the violent tendencies of certain pieces of news.

Furthermore, Wang et al. \cite{wang2023cross} suggested a multimodal fake news detection method using cross-modal contrastive learning. They also noted that multimodal fake news detection models can outperform unimodal baselines. Chen et al. \cite{chen2023multi} also examined multimodal approaches to fake news detection with respect to robustness. However, there are still limitations regarding multilingual fake news detection and sarcasm-aware models. Unlike those, the multimodal framework developed in this chapter uses multilingual text and geospatial metadata.

\subsection{Geospatial Awareness in Recognizing False Information}
Many existing studies that focus on identifying fake news primarily consider text and images, neglecting the role played by spatial information. However, some recent studies have emphasized the importance of spatial cues. Jing et al. \cite{jing2023multimodal} developed a method that leverages spatial and textual information to identify the presence of spatial metadata that helps recognize the spread of misinformation and predict future flashpoints before they cause societal unrest.

Alam et al. \cite{alam2021survey} also emphasized the potential value of geospatial signals in identifying misinformation hotspots and tracking the spread of misleading narratives. Wang et al. \cite{wang2023cross} further observed that regional context can improve fake news identification by revealing local tendencies in misinformation dissemination. Nevertheless, even recent work in low-resource language settings, such as MMCFND \cite{bansal2024mmcfnd} and MMHFND \cite{sharma2024mmhfnd}, remains language-specific and does not incorporate violence-aware geospatial modelling.

\section{Materials and Methods}

\subsection{Dataset}
The proposed dataset combines four major sources: BanFakeNews (50,000 Bangla articles) \cite{zobaer2020banfakenews}, Kaggle Fake News (45,000 English articles) \cite{kagglefake}, Sarcasm Headlines (30,000 headlines) \cite{misra2023sarcasm}, and Twitter Multimodal Fake News (13,000 posts containing text, images, and geotags) \cite{ojo2025dataset}. A standardized format consisting of headline, content, image URL, label, sarcasm, latitude, longitude, and timestamp was applied across all datasets. The merged collection contains 138,256 Bangla and English records. Statistics on the integrated fusion corpus are presented in Table~\ref{tab:dataset}.

The fused corpus is heterogeneous rather than uniformly multimodal. BanFakeNews, Kaggle Fake News, and Sarcasm Headlines contribute text-only records, whereas the Twitter subset contributes records with image references in addition to text. To preserve a unified model input format, missing image inputs were treated as unavailable visual modality and replaced during training with a placeholder image. Similarly, sarcasm was retained as an auxiliary stylistic signal. Because the four source datasets differ in both modality coverage and label semantics, the fused dataset should be interpreted as a heterogeneous benchmark for multimodal-aware misinformation analysis rather than a fully image-complete multimodal corpus.

Because the source datasets differ in annotation purpose and label semantics, a unified binary training space was constructed for model development. The original fake/real labels from the misinformation datasets were retained as the core veracity signal, while the sarcasm dataset was incorporated as a stylistic component and its sarcasm annotation was also preserved separately through the metadata field. This design allows the framework to capture rhetorical cues that may co-occur with misleading or provocative content, while also acknowledging that sarcasm and factual veracity are not identical constructs.

\begin{table}[!htb]
\centering
\caption{Statistics of datasets integrated into the fusion corpus.}
\label{tab:dataset}
\begin{adjustbox}{max width=\linewidth}
\begin{tabular}{lcccc}
\toprule
\textbf{Dataset} & \textbf{Records} & \textbf{Language} & \textbf{Modalities} & \textbf{Labels} \\
\midrule
BanFakeNews \cite{zobaer2020banfakenews} & 49,977 & Bangla & Text & Real/Fake \\
Fake News Detection \cite{kagglefake} & 44,898 & English & Text & Real/Fake \\
Sarcasm Dataset \cite{misra2023sarcasm} & 28,619 & English & Text & Sarcastic/Non-sarcastic \\
Twitter and Weibo \cite{ojo2025dataset} & 14,762 & English & Text+Image & Real/Fake \\
\textbf{Fusion Dataset} & 138,256 & Multilingual & Text + Partial Image + Metadata & Unified Binary Label + Style Metadata \\
\bottomrule
\end{tabular}
\end{adjustbox}
\end{table}

\begin{figure}[!htb]
\centering
\includegraphics[width=\linewidth]{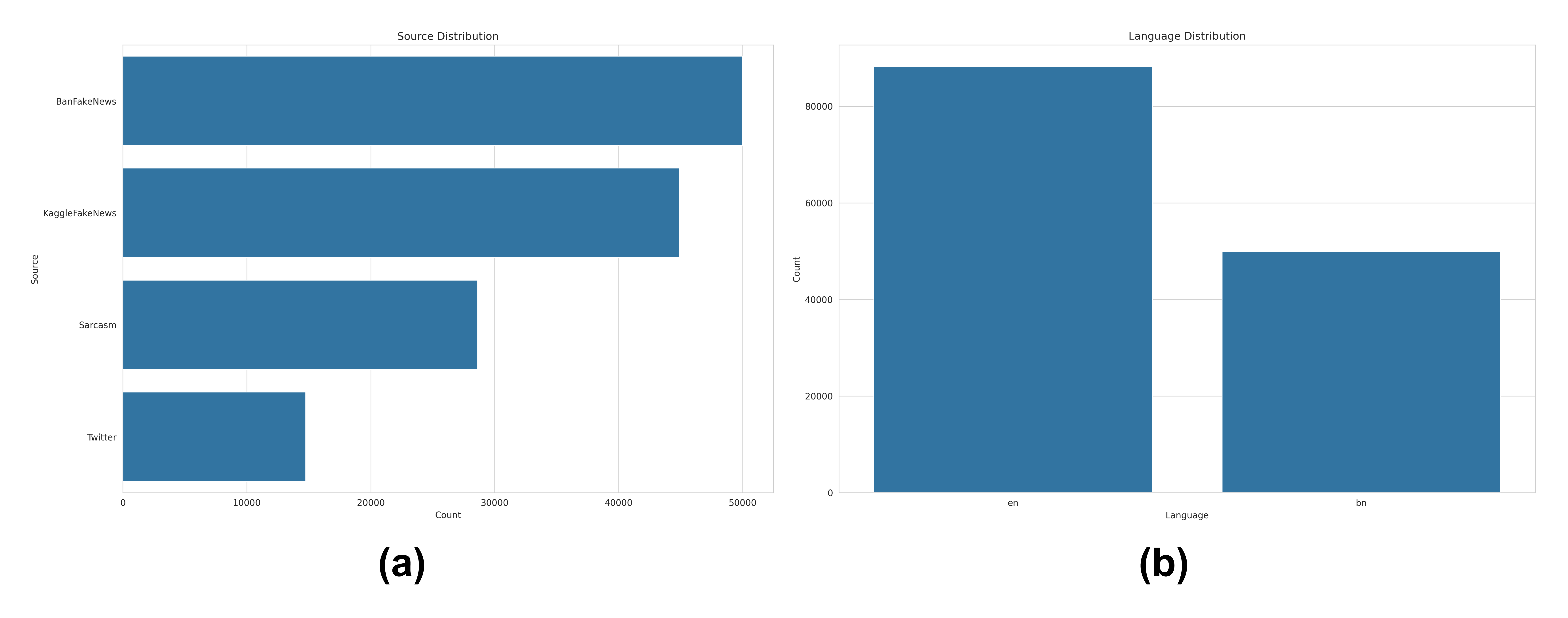}
\caption{Distribution of samples across the four integrated data sources and the two languages represented in the fused corpus.}
\label{fig:dataset_distribution_language_source}
\end{figure}

\begin{figure}[!htb]
\centering
\includegraphics[width=\linewidth]{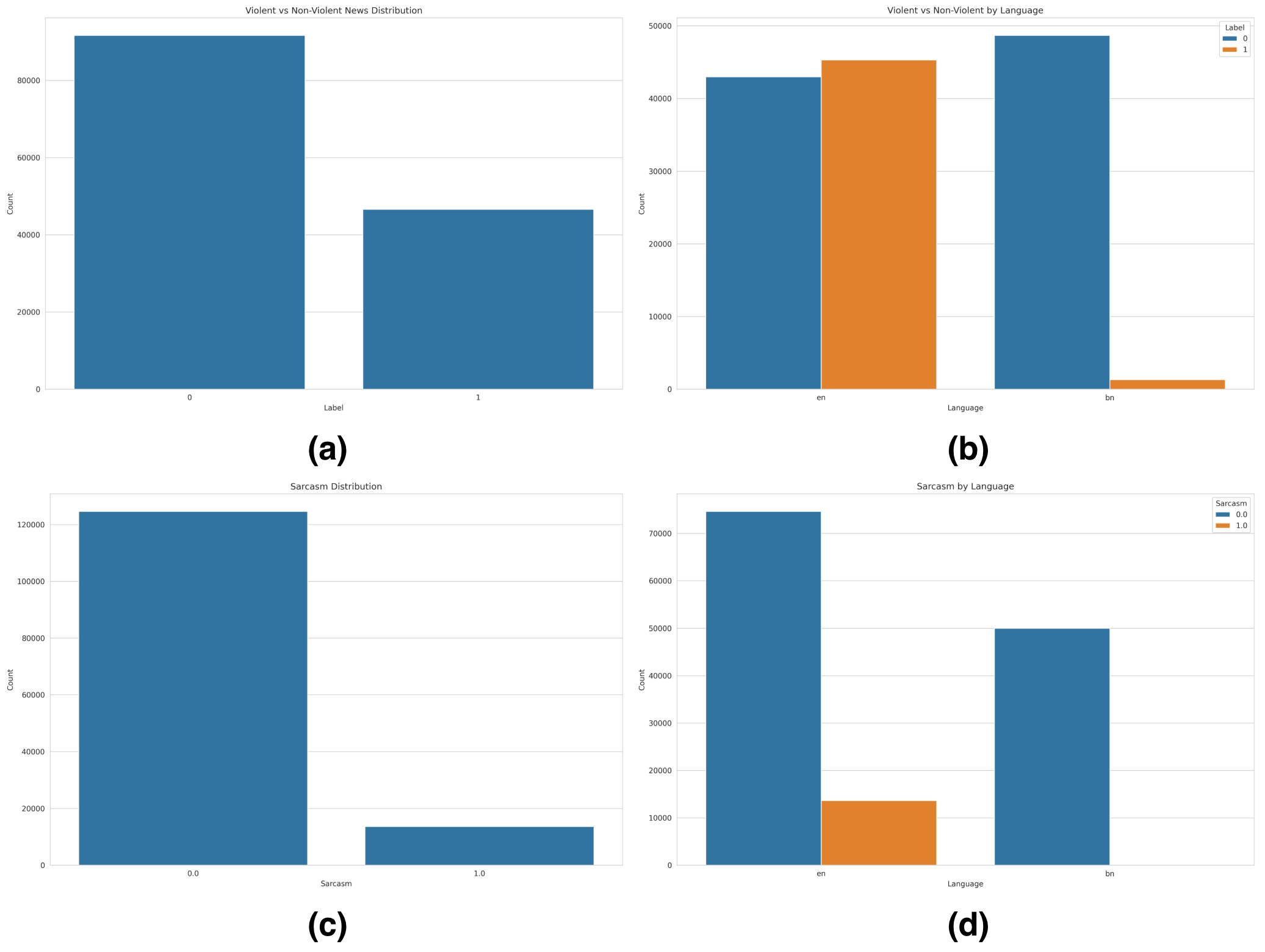}
\caption{Distribution of class labels and sarcasm annotations across the fused multilingual corpus.}
\label{fig:class_distribution}
\end{figure}

Figure~\ref{fig:dataset_distribution_language_source} presents the distribution of samples across data sources and languages in the fused corpus. Figure~\ref{fig:class_distribution} illustrates the distribution of class labels and sarcasm indicators across the benchmark, highlighting the multilingual and heterogeneous nature of the collection.

\begin{figure}[!htb]
\centering
\includegraphics[width=\linewidth]{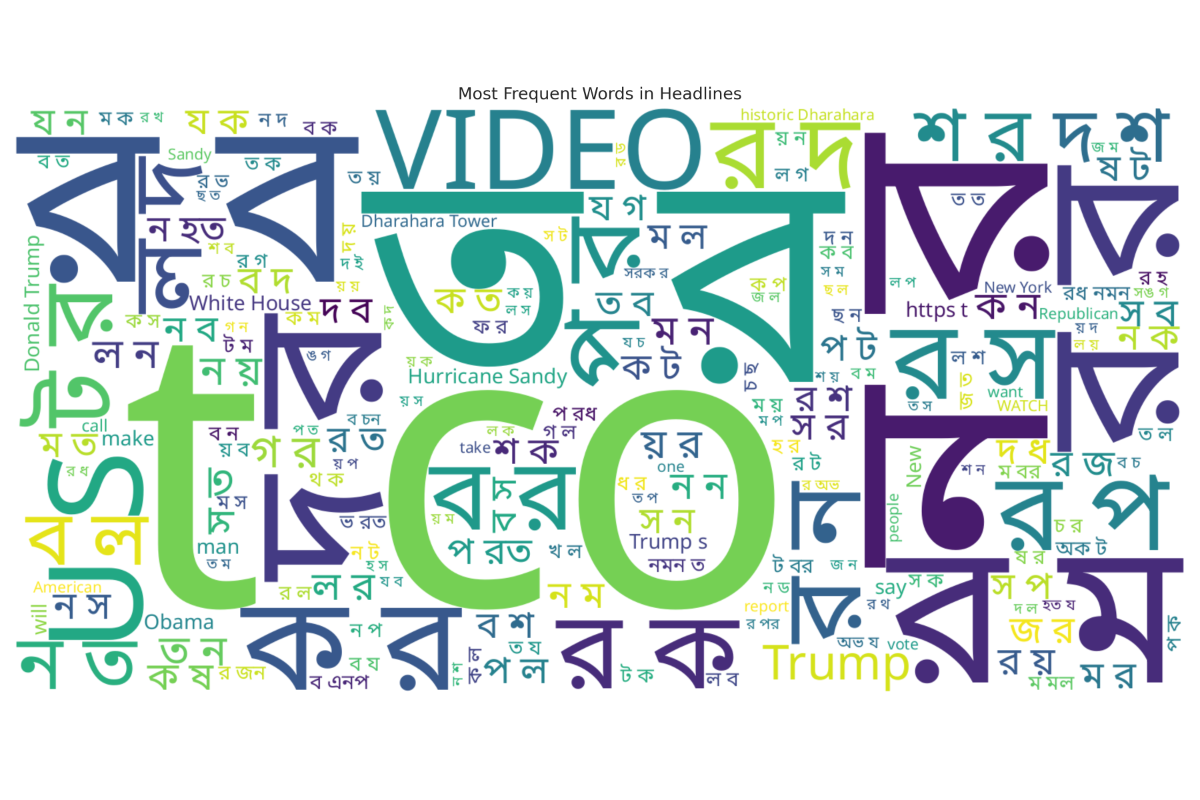}
\caption{Word cloud of the most frequent words in the headlines of the fused multilingual dataset.}
\label{fig:word_frequency}
\end{figure}

Figure~\ref{fig:word_frequency} displays a word cloud created from the most common terms found in the fused dataset's headlines. The visualization highlights commonly occurring terms from both Bangla and English content, reflecting the multilingual nature of the corpus.

\subsection{User Analysis and Design Considerations}
Journalists, non-governmental organizations, security agencies, and policymakers are among the target users of the proposed system. Policymakers require early warnings to plan timely interventions, while hotspot maps may help security personnel focus resources on areas susceptible to instability. Fact-checkers and journalists need to identify false information rapidly in order to ensure reporting accuracy. Therefore, real-time interpretability, multilingual processing capability, and multimodal integration were central design considerations in the proposed framework.

\subsection{Proposed Method}
\begin{figure}[!htb]
\centering
\includegraphics[width=0.85\linewidth]{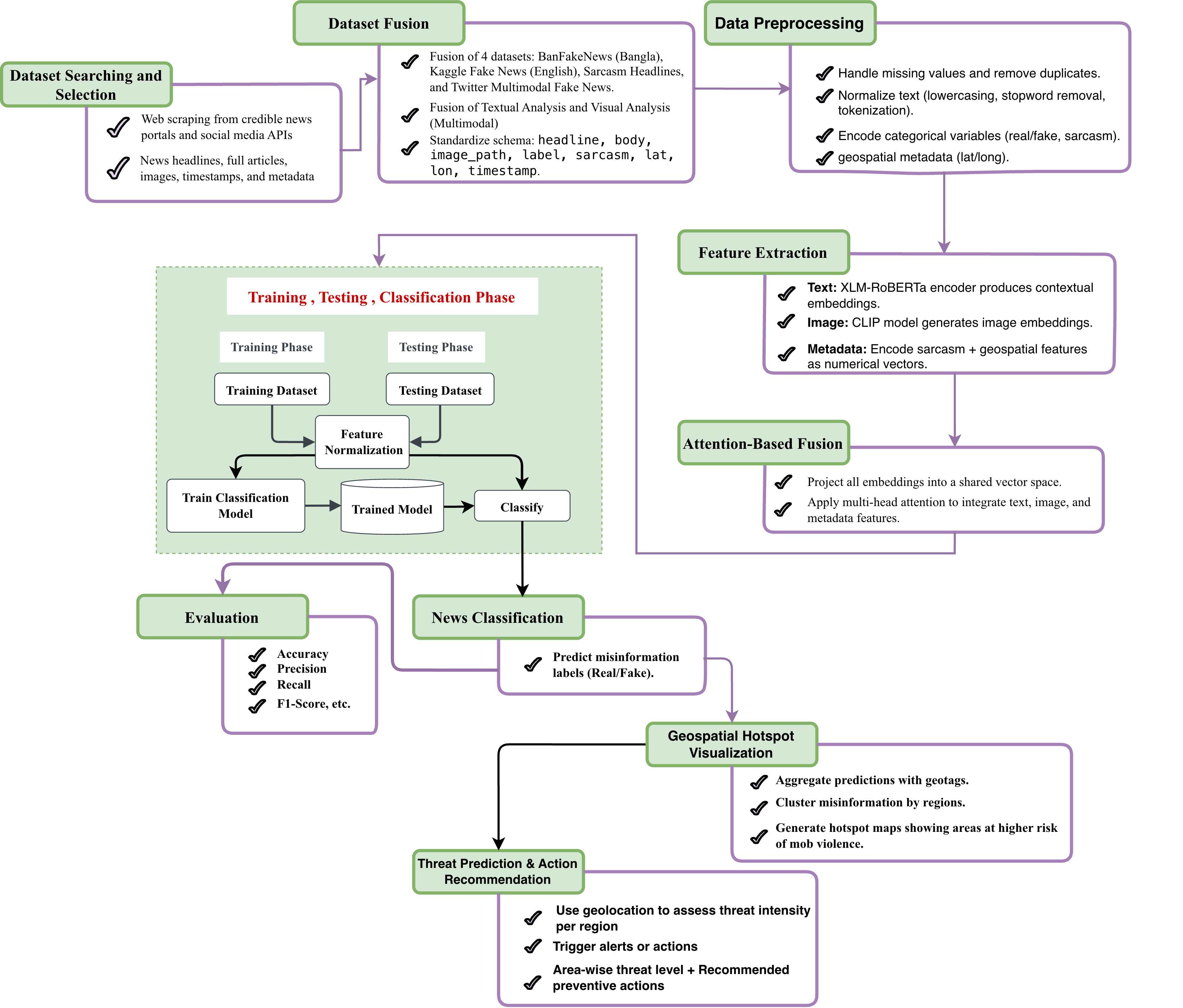}
\caption{Workflow of the proposed multimodal misinformation detection framework.}
\label{fig:workflow}
\end{figure}

The suggested approach combines linguistic, visual, stylistic, and geospatial information to identify misleading claims and monitor their potential escalation into violence. Figure~\ref{fig:workflow} illustrates the six main stages of the framework: integrating data sources, preprocessing, feature extraction, projection into a common latent space, attention-based fusion and classification, and geospatial hotspot visualization.

A combination of BanFakeNews \cite{zobaer2020banfakenews}, Kaggle Fake News \cite{kagglefake}, Sarcasm Headlines \cite{misra2023sarcasm}, and multimodal tweets \cite{ojo2025dataset} was merged into one unified collection containing text, images, sarcasm cues, and geospatial details. Before feature extraction, the data were standardized through preprocessing, including text normalization, image resizing, and metadata encoding. XLM-RoBERTa was used to process text, CLIP to extract visual representations, and numerical encoding to represent metadata. These features were projected into a common latent space and combined through an attention-based fusion mechanism that dynamically weighed multiple cues. A dense layer then generated real or fake predictions using a softmax function. These predictions were subsequently aggregated by location and visualized as hotspot maps in order to provide early warning signals. The full process for misinformation detection and unrest monitoring is summarized in Table~\ref{alg:multimodal}.

\begin{table}[!htb]
\centering
\caption{Proposed attention-based multimodal fusion framework for fake news detection and violence monitoring.}
\label{alg:multimodal}
\begin{adjustbox}{max width=\linewidth}
\begin{tabular}{p{0.94\linewidth}}
\toprule
\textbf{Input:} Dataset $D=\{T,I,M\}$, where $T$ denotes text, $I$ denotes image, and $M=[s,g]$ includes sarcasm flag $s$ and geospatial coordinates $g$. \\
\textbf{Output:} Predicted label $\hat{y}\in\{0,1\}$ (Real/Fake), with optional geospatial hotspot map. \\
\midrule
\textbf{Step 1: Preprocessing.} Normalize and tokenize $T$; resize and normalize $I$; encode sarcasm flag $s$; clean geospatial metadata $g$. \\
\textbf{Step 2: Feature Extraction.} Compute $X_t=f_{\text{text}}(T)$ using XLM-RoBERTa; compute $X_i=f_{\text{image}}(I)$ using CLIP; compute $X_m=f_{\text{meta}}([s,g])$. \\
\textbf{Step 3: Projection.} $H_t=W_tX_t$, $H_i=W_iX_i$, $H_m=W_mX_m$. \\
\textbf{Step 4: Attention-Based Fusion.} $Z=\text{Attention}([H_t,H_i,H_m])$. \\
\textbf{Step 5: Classification.} $\hat{y}=\text{Softmax}(WZ+b)$. \\
\textbf{Step 6: Optional Geospatial Visualization.}  When valid coordinates are available, Aggregate $\{g \mid \hat{y}=1\}$, cluster results, and visualize misinformation intensity. \\
\bottomrule
\end{tabular}
\end{adjustbox}
\end{table}

\subsection{Architecture Development}
\begin{figure}[!htb]
\centering
\includegraphics[width=0.80\linewidth]{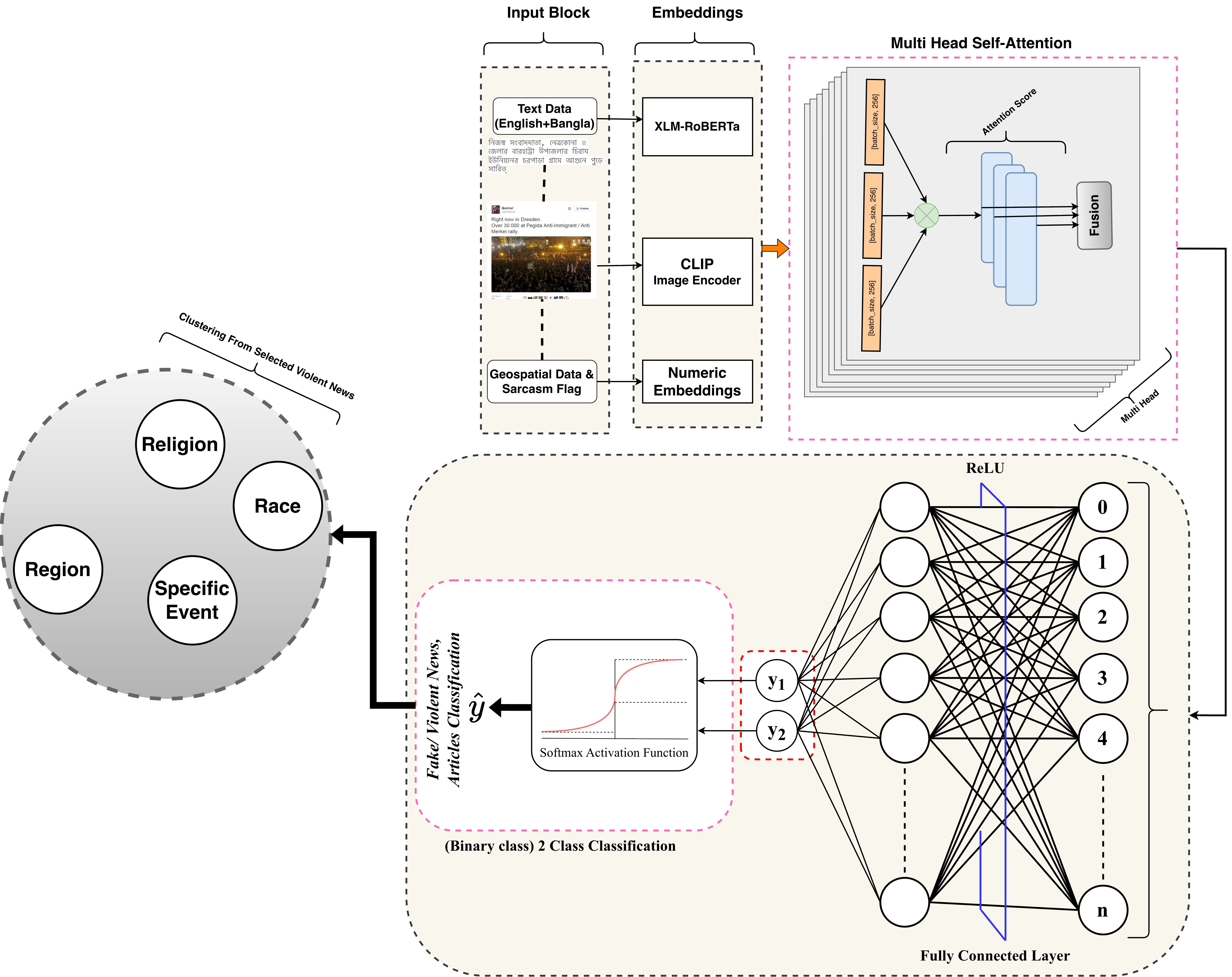}
\caption{Proposed architecture of the attention-based fusion model.}
\label{fig:arch}
\end{figure}

Figure~\ref{fig:arch} shows the structure of the AttentionFusion model. Instead of treating inputs separately, the framework combines text, images, and metadata such as geolocation and sarcasm indicators. Text written in either English or Bangla is processed using XLM-RoBERTa, while images are passed through a CLIP-based feature extractor. Metadata are converted into numerical form and projected into a shared hidden space through a dense neural layer. These representations are then fused through a multi-head self-attention mechanism that dynamically adjusts feature importance across modalities.

For records without available images, a placeholder visual input was used during preprocessing so that a consistent multimodal interface could be maintained across heterogeneous data sources.

After fusion, the integrated representation is processed through a dense neural network with ReLU activations and a final softmax classification layer trained to distinguish fake and genuine news. When coordinate metadata are available and sufficiently informative, predicted misinformation cases may also be aggregated spatially to support exploratory hotspot analysis.

The architectural configuration of the proposed AttentionFusion model, including output dimensions and trainable parameters of its main components, is shown in Table~\ref{tab:architecture_parameters}.

\begin{table}[!htb]
\centering
\caption{Architectural parameter configuration of the proposed AttentionFusion model.}
\label{tab:architecture_parameters}
\begin{adjustbox}{max width=\linewidth}
\begin{tabular}{lcc}
\toprule
\textbf{Layer} & \textbf{Output Shape} & \textbf{Trainable Parameters} \\
\midrule
XLM-RoBERTa & (batch, 768) & 278043648 \\
CLIP Image Encoder & (batch, 512) & 11176512 \\
Metadata FC & (batch, 256) & 1024 \\
Text Projection & (batch, 256) & 196864 \\
Image Projection & (batch, 256) & 131328 \\
Multi-head Attention & (3, batch, 256) & 263168 \\
Classifier & (batch, 2) & 66306 \\
\bottomrule
\end{tabular}
\end{adjustbox}
\end{table}

\section{Experimental Results and Discussion}

\subsection{Experimental Setup}
To evaluate the effectiveness of the suggested attention-based fusion framework, experiments were conducted on a stratified random sample comprising 30\% of the fused dataset. From the original 138,256 records, 41,477 samples were used for model development. This sampling strategy was adopted to maintain computational feasibility while preserving class diversity and multilingual coverage across the integrated benchmark.

The experiment was conducted with a fixed random seed of 42. There were 29,033 training instances, 6,222 validation instances, and 6,222 test instances after the sampled dataset was split into training, validation, and test subsets in a stratified 70:15:15 ratio. XLM-RoBERTa was used to tokenize textual inputs with a maximum sequence length of 128. For 10 epochs, the model was trained with a batch size of 8 and a learning rate of $2 \times 10^{-5}$
, a dropout of 0.2, a hidden dimension of 256, and the Adam optimizer.

Visual features were extracted using CLIP and combined with textual and metadata features through the proposed attention-based fusion mechanism. Metadata were represented through sarcasm, latitude, and longitude fields. When an image was unavailable, a placeholder visual input was used to preserve a consistent model interface across heterogeneous source datasets. Performance was assessed using a confusion matrix, precision, recall, F1-score, ROC curve, AUC, and training-validation accuracy and loss curves.

Because the fused benchmark integrates heterogeneous sources with different annotation purposes and modality coverage, the reported results should be interpreted as a benchmark evaluation on the constructed corpus rather than as a claim of universal real-world generalization.

\subsection{Performance Evaluation}
\begin{figure}[!htb]
\centering
\includegraphics[width=\linewidth]{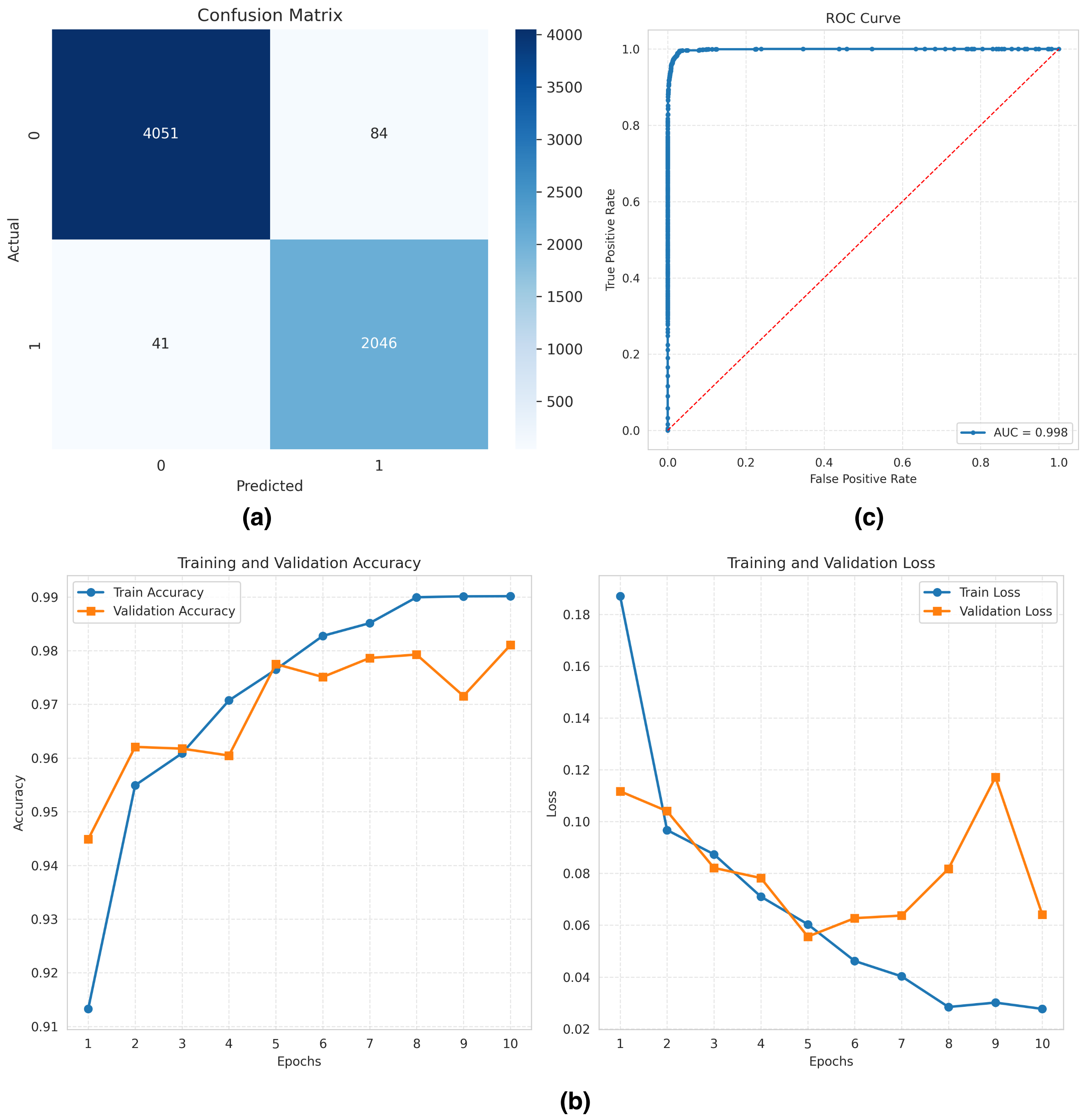}
\caption{Performance evaluation of the proposed multimodal fusion framework for detecting fake news and violence-driven mob activity: (a) confusion matrix showing class-wise prediction outcomes, (b) training and validation accuracy and loss curves over 10 epochs, and (c) ROC curve with an AUC of 0.998, demonstrating strong discriminative ability and stable convergence.}
\label{fig:cm_loss_accuracy_curve}
\end{figure}

As shown in Figure~\ref{fig:cm_loss_accuracy_curve}(a), the confusion matrix provides a clear demonstration of the predictive dynamics of the model on a class-by-class basis. It correctly identified 4,051 Class~0 instances and 2,046 Class~1 instances. Meanwhile, 84 instances of Class~0 were incorrectly identified as Class~1 and 41 instances of Class~1 were incorrectly identified as Class~0. The small number of misclassifications in both categories indicates that the proposed framework is highly consistent and reliable in distinguishing between fake news and violence-motivated mob behaviour.

A more detailed quantitative evaluation is presented in Table~\ref{tab:classification_report}. For Class~0, the model obtained a precision of 0.99, a recall of 0.98, and an F1-score of 0.98 on 4,135 test samples. For Class~1, the respective values were 0.96, 0.98, and 0.97 on 2,087 samples. The overall test accuracy was 0.98, and both the macro-average F1-score and weighted-average F1-score were approximately 0.98. These results show that the model has robust and balanced predictive accuracy across both classes, rather than merely favoring the dominant category.
\begin{table}[!htb]
\centering
\caption{Classification report of the proposed model on the test set.}
\label{tab:classification_report}
\begin{adjustbox}{max width=0.9\linewidth}
\begin{tabular}{lcccc}
\toprule
\textbf{Class} & \textbf{Precision} & \textbf{Recall} & \textbf{F1-Score} & \textbf{Support} \\
\midrule
Class 0 (Real) & 0.99 & 0.98 & 0.98 & 4135 \\
Class 1 (Fake) & 0.96 & 0.98 & 0.97 & 2087 \\
\midrule
\textbf{Accuracy} & 0.98 & 0.98 & 0.98 & 0.98 \\
\textbf{Macro Avg.} & 0.97 & 0.98 & 0.97 & 6222 \\
\textbf{Weighted Avg.} & 0.98 & 0.98 & 0.98 & 6222 \\
\bottomrule
\end{tabular}
\end{adjustbox}
\end{table}

Although the overall training process was stable, the fluctuation in validation loss during later epochs suggests the possibility of mild overfitting after mid-training. This behaviour did not substantially reduce held-out test performance, but it indicates that future versions of the framework would benefit from stricter regularization, early stopping, and additional robustness checks. In addition, the present study evaluates a single fusion architecture; future work should include direct comparison with text-only, text-plus-metadata, and alternative multimodal baselines in order to more clearly quantify the advantage of the suggested model.

The discriminative strength of the proposed framework is further supported by the ROC curve shown in Figure~\ref{fig:cm_loss_accuracy_curve}(c), where the model achieved an area under the curve (AUC) of approximately 0.99. This high AUC value indicates excellent separability between the two classes across a wide range of threshold settings, confirming the robustness of the classifier beyond a single operating point.

Figure~\ref{fig:cm_loss_accuracy_curve}(b) shows the training and validation accuracy and loss curves, providing additional insight into the model’s training behaviour. Validation accuracy increased from 94.49\% to 98.10\%, while training accuracy rose from 91.33\% to 99.01\% over 10 epochs. Simultaneously, training loss decreased from 0.1871 to 0.0277, and validation loss generally declined from 0.1117 to 0.0641. These trends indicate stable convergence and effective parameter optimization throughout training.

Overall, the experimental findings show the effectiveness of the suggested multimodal framework and validate its suitability for accurately identifying fake news and mob activity motivated by violence. Comparative performance against recent related studies is presented in Table~\ref{tab:stateart}.

\begin{table}[!htb]
\centering
\caption{State-of-the-art comparative analysis of the proposed method with recent related studies}
\label{tab:stateart}

\small
\setlength{\tabcolsep}{4pt}
\renewcommand{\arraystretch}{1.15}

\begin{adjustbox}{max width=\textwidth}
\begin{tabular}{llllll}
\toprule
\textbf{Study} & \textbf{Language} & \textbf{Modalities} & \textbf{Method} & \textbf{Accuracy} & \textbf{Limitations} \\
\midrule

Ref.~\cite{singh2021detecting} & English & Text + Visual & Multimodal Analysis & 85.25\% & English-only \\

Ref.~\cite{qi2021improving} & Chinese+English & Text + Visual & Entity Fusion & up to 97.5\% & Weak sarcasm cues \\

Ref.~\cite{xu2025mdam3} & English & Text + Image + Video & MDAM3 & 85.3\% & High compute cost \\

Ref.~\cite{wang2023cross} & English & Text + Visual & Robustness Eval & up to 92.5\% & Adversarial fragility \\

Ref.~\cite{bansal2024mmcfnd} & Indic & Text + Visual & Caption-Aware & 99.6\% & Low-resource \\

Ref.~\cite{sharma2024mmhfnd} & Hindi & Text + Visual & Contrastive Learning & $\sim$98.6\% & Language-specific \\

\textbf{Proposed Model} & Bangla+English+Meta & Text+Image+Sarcasm+Geo & Fusion Model & 98.0\% & Violence-aware \\

\bottomrule
\end{tabular}
\end{adjustbox}

\normalsize
\end{table}

\FloatBarrier
\section{Conclusion and Future Work}
In this chapter, we present a model developed through multi-language and multi-modal inputs that is used to detect misinformation likely to instigate violence. The model achieves accurate classification by using textual information, visual information, sarcasm, and geographical data. One of the major advantages of this model is that it can help map misinformation hot spots, which makes it applicable for law enforcement officers, media personnel, and policymakers in locating areas where there is a higher probability of violence.

Despite the encouraging outcomes, there are still a few restrictions. First, the merged dataset contains varied content, but only a portion of it is accompanied by images, making it difficult to fully explore the multimodal aspects of the dataset. Secondly, each individual source dataset has different purposes for its labels, including sarcasm. Third, the experiments were conducted on a stratified 30\% sample of the corpus for computational reasons rather than on the full dataset. Fourth, the present evaluation focuses on internal benchmark performance and does not yet include source-isolated validation, external real-world testing, or formal statistical significance analysis. These limitations mean that the reported results should be interpreted as encouraging but still preliminary.

Future work will focus on constructing more fully aligned fused datasets, evaluating the framework under stricter split strategies, and comparing it against stronger unimodal and multimodal baselines. Extending the framework to low-resource and underrepresented languages remains a major priority. Incorporating additional modalities such as audio and video, together with more reliable location-aware metadata and broader social media signals, may further strengthen performance. Another important direction is the deployment of scalable real-time monitoring pipelines and interactive dashboards. With these extensions, the proposed framework may evolve into a more comprehensive misinformation tracking system for early detection and intervention.

\FloatBarrier